\documentclass[sigconf]{acmart}

\usepackage[most]{tcolorbox}

\usepackage[normalem]{ulem}
\usepackage{tabularray}
\usepackage{makecell} 
\usepackage[normalem]{ulem}
\useunder{\uline}{\ul}{}
\usepackage{algorithm}
\usepackage{algorithmic}
\usepackage{multirow}
\usepackage{amsmath} 
\usepackage{enumitem}
\usepackage{balance}
\usepackage{caption} 

\usepackage{array} 
\usepackage{etoolbox} 
\usepackage{graphicx}
\usepackage{booktabs}

\usepackage{hyperref}
\usepackage{url}
\usepackage{amsmath} 
\usepackage{subcaption}
\usepackage{tcolorbox}
\usepackage{fancyvrb}
\usepackage{enumitem}
\usepackage{booktabs}       
\usepackage{amsfonts}       
\usepackage{nicefrac}       
\usepackage{microtype}      
\usepackage{xcolor}         
\usepackage{graphicx}
\usepackage{tcolorbox}
\usepackage{caption}

\copyrightyear{2026}
\acmYear{2026}
\setcopyright{cc}
\setcctype{by}
\acmConference[SIGIR '26]{Proceedings of the 49th International ACM SIGIR Conference on Research and Development in Information Retrieval}{July 20--24, 2026}{Melbourne, VIC, Australia}
\acmBooktitle{Proceedings of the 49th International ACM SIGIR Conference on Research and Development in Information Retrieval (SIGIR '26), July 20--24, 2026, Melbourne, VIC, Australia}
\acmDOI{10.1145/3805712.3808598}
\acmISBN{979-8-4007-2599-9/2026/07}

\begin{document}
\title{SurGE: A Benchmark and Evaluation Framework for Scientific Survey Generation}

\author{Weihang Su}
\email{swh22@mails.tsinghua.edu.cn}
\affiliation{%
    \institution{Tsinghua University}
    \city{Beijing}
    \country{China}
}
\affiliation{%
    \institution{Quan Cheng Laboratory}
    \city{Shandong}
    \country{China}
}

\author{Anzhe Xie}
\authornote{Contributed equally}
\affiliation{%
    \institution{Tsinghua University}
    \city{Beijing}
    \country{China}
}

\author{Qingyao Ai}
\authornote{Corresponding author}
\email{aiqy@tsinghua.edu.cn}
\affiliation{%
    \institution{Quan Cheng Laboratory}
    \city{Shandong}
    \country{China}
}
\affiliation{%
    \institution{Tsinghua University}
    \city{Beijing}
    \country{China}
}

\author{Jianming Long}
\affiliation{%
    \institution{Tsinghua University}
    \city{Beijing}
    \country{China}
}

\author{Xuanyi Chen}
\affiliation{%
    \institution{Tsinghua University}
    \city{Beijing}
    \country{China}
}

\author{Jiaxin Mao}
\affiliation{%
    \institution{Renmin University of China}
    \city{Beijing}
    \country{China}
}

\author{Ziyi Ye}
\affiliation{%
    \institution{Fudan University}
    \city{Beijing}
    \country{China}
}

\author{Yiqun Liu}
\affiliation{%
    \institution{Tsinghua University}
    \city{Beijing}
    \country{China}
}

\begin{abstract}
The exponential growth of scientific literature has created a pressing need for automated survey generation. Although recent LLM-based agents have shown promise in automating this task, current progress is hindered by the lack of a standardized, scalable evaluation protocol. Existing evaluation methods typically rely on either human evaluation or custom metrics designed to validate specific pipelines, which restricts scalability and hinders fair comparison. To address this, we introduce \textbf{SurGE}, a benchmark and evaluation framework tailored for scientific survey generation. SurGE provides a large-scale retrieval corpus of over one million papers and expert-validated ground-truth surveys. Furthermore, we propose a robust multi-dimensional evaluation protocol that integrates both objective metrics and LLM-based judgments, and empirically verify its high alignment with human experts. Our experiments reveal that while agentic pipelines outperform RAG baselines in fluency and structural quality, they still struggle with citation accuracy, highlighting key directions for future research\footnote{We have open-sourced all the code and data at: \url{https://github.com/oneal2000/SurGE}}. 
\end{abstract}


\keywords{Scientific Survey Generation, Benchmark, Evaluation Framework, Large Language Models, Retrieval-Augmented Generation}

\ccsdesc[500]{Information systems~Information retrieval}

\maketitle

\section{Introduction}
\label{sec:1}

The volume of scientific literature has been expanding at an unprecedented rate in recent years. 
For instance, arXiv now receives over a thousand new computer science papers daily, which is more than doubled between 2019 and 2024~\citep{liang2025surveyx}.
This rapid growth in publications has made the manual creation of comprehensive survey papers increasingly impractical,
as collecting and synthesizing such vast volumes of information is both labor-intensive and time-consuming. Consequently, there is a critical need for automated systems capable of generating high-quality academic surveys.

Recent LLM-based agents have begun to address this need by automating survey generation via hierarchical decomposition and memory-driven mechanisms~\citep{wang2024autosurvey,yan2025surveyforge,lai2024instruct}.
However, while these systems establish the feasibility of automated survey writing, the field still lacks a standardized evaluation protocol.
Existing evaluation practices are either unscalable or tightly coupled to specific generation pipelines.
For example, StepSurvey relies heavily on human evaluation~\citep{nlpcc2024}, limiting scalability and reproducibility.
In contrast, AutoSurvey and SurveyForge introduce customized evaluation frameworks primarily designed to validate their own pipelines.
While effective for internal analysis, such evaluations are inherently pipeline-dependent, leading to self-validation that favors design-specific assumptions.
This dependency introduces systematic bias and prevents fair, cross-system comparison.
As the community shifts from demonstrating the feasibility of survey generation to rigorously assessing its quality, a method-agnostic, reproducible, and diagnostic benchmark becomes essential.
Such a benchmark should disentangle the ability to retrieve relevant literature from the ability to synthesize it into a coherent survey. This decomposition is crucial for diagnostic evaluation, as errors in the final survey may originate from either stage.

To address this gap, we introduce \textbf{SurGE} (\underline{Sur}vey \underline{G}eneration \underline{E}valuation), 
a comprehensive benchmark and evaluation framework for scalable, fair, and reproducible assessment of survey generation systems.
We formalize survey generation as a two-stage process: (1) retrieving relevant papers from a comprehensive corpus, and (2) synthesizing the retrieved evidence into a structured survey. 
To facilitate this, SurGE provides a rigorously constructed dataset comprising a large-scale academic corpus (over one million papers) and a collection of high-quality test instances with ground truth. 
Ground-truth surveys are sourced from high-impact, peer-reviewed publications and are further validated by expert annotators to ensure high reliability.

Beyond dataset construction, a key contribution of SurGE is its robust, multi-dimensional evaluation framework specifically tailored for survey generation. 
Grounded in established principles for high-quality survey writing~\citep{webster2002analyzing}, our framework combines objective, quantifiable indicators (e.g., citation coverage) with LLM-as-a-Judge metrics (e.g., logical coherence and content quality).
Crucially, to ensure the reliability of our proposed evaluation framework, we conduct a rigorous meta-evaluation: we collect fine-grained expert annotations and verify a high correlation between human judgment and our LLM-based metrics. 
This alignment confirms that SurGE is a reliable and scalable proxy for evaluating AI-generated surveys.

To empirically assess the utility of {SurGE}, we benchmark a wide range of LLM-based systems. 
Leveraging the decoupled retrieval-generation design of our framework, we conduct an in-depth analysis to disentangle whether performance bottlenecks arise from information acquisition or content synthesis. 
Our experiments demonstrate that specialized agentic pipelines consistently outperform standard RAG baselines. However, while these agents exhibit superior linguistic fluency, they struggle with information utilization, failing to fully leverage the collected evidence during the generation phase.
In summary, our contributions are threefold:

\begin{enumerate}[leftmargin=*,itemsep=0pt] 

\item We introduce \textbf{SurGE}, a method-agnostic benchmark for survey generation featuring expert-validated ground-truth surveys and a large-scale retrieval corpus. 

\item We propose an automated evaluation framework that assesses survey quality across four crucial dimensions: comprehensiveness, citation accuracy, structure, and content.

\item We conduct extensive experiments on representative state-of-the-art systems, pointing out directions for future optimization.

\end{enumerate}

\section{Task Definition}
\label{sec:Task_Definition}

We formalize survey generation as a two-stage task. 
Given a topic description $t$ and an academic corpus $D = \{d_1, d_2, \ldots,d_n\}$, the goal is to generate a survey article $S$ that provides a structured and comprehensive overview of the topic.
The process consists of:

\begin{itemize}[leftmargin=*]
    \item \textbf{Information Collection:} A retrieval system collects a relevant paper set $\mathcal{R}_t \subseteq D$ containing papers relevant to $t$.
    
    \item \textbf{Survey Generation:} A generative system composes a survey $S$ based on the topic $t$ and the retrieved document set $\mathcal{R}_t$, including proper citations and a proper reference list.
    
\end{itemize}

\section{Dataset Construction}
\label{sec:3}
This section presents the construction process of the SurGE benchmark. 
We first describe how we collect high-quality ground-truth surveys in Section~\ref{sec:selection}, followed by the expert annotation protocol in Section~\ref{sec:annotation}. 
We then introduce the construction of the large-scale academic corpus in Section~\ref{sec:corpus}, and report dataset statistics and analysis in Section~\ref{sec:statistics}. 
Finally, we discuss the ethical and licensing considerations underlying our data collection protocol in Section~\ref{sec:ethics}.

\subsection{Ground-Truth Survey Collection}
\label{sec:selection}

To construct the SurGE benchmark, we began by collecting a diverse set of high-quality reference surveys from recent computer science literature. The candidate papers were sourced from the arXiv repository, focusing on publications between 2020 and 2024 that explicitly identified themselves as survey articles or systematic reviews. We applied the following selection criteria to ensure academic significance and reliability: (i) the document must explicitly declare itself as a survey or review; (ii) it must have a minimum citation count of 20 to indicate scholarly impact~\citep{bornmann2008citation}; (iii) the publication date must fall within the range of 2020 to 2024.

\subsection{Expert Annotation}
\label{sec:annotation}
Building on the initial automated collection, we implemented a secondary manual filtering step to ensure the dataset's quality. 
To this end, we recruited a team of four Ph.D. students in computer science, led by a faculty member, to serve as annotators.
Each candidate document was evaluated by two independent annotators along four key dimensions: (i) citation impact, reflecting the scholarly influence of the paper; (ii) content coverage, indicating how comprehensively the survey summarizes the literature within its scope; (iii) structural coherence, assessing the logical organization and clarity of the document’s sections; and (iv) citation quality, which examines the relevance, diversity, and traceability of cited works.

Annotation averaged 10 minutes per paper.
Each annotator labeled the document as either ``usable'' or ``not usable''. 
Papers were retained only if labeled as ``usable'' by both annotators, discarding any disagreements to maintain a high quality threshold. Annotators were compensated 60 CNY/hour (much exceeding the local minimum wage).
Inter-annotator agreement was quantified using Cohen’s Kappa, applied to 250 annotated instances. 
The resulting score of 0.792 indicates substantial agreement, highlighting the reliability of the quality control process. 
After this filtering stage, we finalized the dataset with 205 rigorously verified survey papers~\footnote{The final set of 205 ground-truth surveys included in our benchmark is available in our official repository at {\url{https://github.com/oneal2000/SurGE/blob/main/gt_surveys.tex}}}.

\subsection{Academic Corpus Construction}
\label{sec:corpus}

A crucial component of the \textbf{SurGE} benchmark is a large-scale academic corpus that serves as the retrieval pool for the document collection stage. Our corpus is built entirely from scholarly metadata obtained from the arXiv repository. 
To ensure adherence to ethical and legal standards, we exclusively collected metadata and did not include full-text PDFs, a practice permitted by arXiv’s Terms of Use, which designates metadata as public domain under the CC0 license~\citep{arxiv_tou}.
The corpus was constructed through the following process. 
First, we retrieve the arXiv metadata for all publicly accessible papers cited in the ground truth survey.
This process revealed that approximately 30\% of the references were unavailable, primarily because they were published in closed-access journals or other restricted venues. 
Then we expanded the corpus to enhance its comprehensiveness. We queried the official arXiv search API, using keywords and titles from the ground-truth surveys to identify and collect metadata for other topically related papers.
To mitigate topic bias, we augment the corpus with a random sample of arXiv computer science papers, drawn independently of the benchmark topics.
The entire process resulted in a final retrieval corpus of 1,086,992 unique papers. 
For each paper, the corpus provides rich metadata, including the title, authors, abstract, subject categories, publication date, and a direct link to the paper's arXiv page for transparency and verification. 
To ensure high data quality, all collected metadata underwent a rigorous preprocessing pipeline that included text normalization, formatting removal, and deduplication.

\subsection{Statistics and Analysis}
\label{sec:statistics}

The resulting SurGE benchmark comprises 205 ground-truth survey papers and a retrieval corpus of 1,086,992 documents. Table~\ref{tab:dataset_statistics} presents the key statistics of the dataset. 
We model hierarchical headings as trees to quantify the complexity of the ground truth surveys. The surveys exhibit deep structures (avg. depth 3.07, 42.7 nodes), posing significant challenges for hierarchical text generation.
Furthermore, the surveys are densely referenced, citing an average of 65.8 papers, which underscores the demand for high-recall information collection. 

\subsection{Ethical Considerations and Licensing}
\label{sec:ethics}

Our corpus is constructed exclusively from arXiv-provided {descriptive metadata} (titles, authors, abstracts, identifiers, categories, and license URIs) harvested via the official API. We do not host or redistribute arXiv PDFs or source files. This design complies with arXiv’s API Terms of Use, which place descriptive metadata under a CC0 public-domain dedication~\citep{arxiv_tou}. This design is also consistent with the arXiv Submittal Agreement’s CC0 designation for metadata~\citep{arxiv_submission}. 
The dataset and codebase are distributed under the MIT license, granting researchers and developers unrestricted access and modification rights.

\begin{table}[t]
    \centering
    \small 
    \caption{Key statistics of the SurGE benchmark.}
    \label{tab:dataset_statistics} 
    \begin{tabular}{@{}l r@{}} 
    \toprule
    \textbf{Statistic} & \textbf{Number} \\
    \midrule
    Total Ground Truth Surveys        & 205 \\
    Average Tree Depth                & 3.073 \\
    Maximum Tree Depth                & 4 \\
    Average Number of Tree Nodes      & 42.717 \\
    Maximum Number of Tree Nodes      & 212 \\
    Average Citations per Paper       & 65.78 \\
    Corpus Size                       & 1,086,992 \\
    Average Abstract Length (words)   & 156.57 \\
    \bottomrule
    \end{tabular}
\end{table}

\section{Evaluation Framework}
\label{sec:evaluation}
\label{sec:4}

To comprehensively evaluate the quality of automatically generated scientific surveys, we propose a multi-faceted evaluation framework. This framework assesses survey quality across four crucial dimensions: {Comprehensiveness}, {Citation Accuracy}, {Structural Quality}, and {Content Quality}. Each generated survey is evaluated against an expert-written Ground Truth (GT) survey. The following subsections detail the quantitative metrics for each dimension.

\subsection{Comprehensiveness}

The comprehensiveness of a scientific survey is a critical quality factor, as the omission of key publications can undermine its credibility and value. 
To quantify this aspect, we evaluate the Recall of a generated survey's references against the ground-truth reference lists.
Formally, let ${R}_{GT}$ be the set of references in an expert-written GT survey and ${R}_{G}$ be the set of references in our generated survey. Recall $\mathcal{R}$ is defined as:

\begin{equation}
\mathcal{R} = \frac{|{R}_{GT} \cap {R}_{G}|}{|{R}_{GT}|},
\end{equation}

\textbf{Crucially}, while the GT reference set is not assumed to be perfectly complete, it serves as the best available proxy for expert consensus on a topic's core literature, given that our GT surveys are highly cited and peer-reviewed. 
We therefore interpret this metric not as a measure of absolute completeness, but as a pragmatic metric for evaluating a system's ability to identify the central body of work validated by the research community.

\subsection{Citation Accuracy}

Citation accuracy is another critical aspect of a high-quality survey. 
Each citation should be thematically relevant to the overall topic, and it must be contextually appropriate in terms of both the section and sentence in which it appears. 
To evaluate this aspect, we introduce Citation Accuracy, which evaluates each citation in the survey across three levels. 
First, at the document level, we assess whether a cited paper is thematically relevant to the survey's overall topic. 
Second, at the section level, we evaluate whether a citation is placed in a semantically appropriate section of the survey. 
Finally, at the sentence level, we verify whether a citation supports the specific claim made in the sentence where it is cited.

To automate this evaluation, we employ a Natural Language Inference (NLI) model (\texttt{nli-deberta-v3-base}~\cite{he2021debertav3} to assess the relevance of each citation. 
An NLI model is designed to determine the logical relationship between two text snippets: a \textbf{premise} and a \textbf{hypothesis}. The model predicts the relationship between these two components, providing probabilities for the following labels: {\textsc{Entailment}, \textsc{Neutral}, \textsc{Contradiction}}.  Due to its ability to capture semantic relationships, NLI has become a standard method for evaluating the factual consistency of text generated by LLMs~\cite{min2023factscore}.
Specifically, we implement this hierarchical evaluation by framing it as a series of NLI tasks. For each citation $r$ (with title $T_r$ and abstract $A_r$) within the generated survey $S$ (with title $T_S$), we construct a set of premise-hypothesis pairs. 
The premise is consistently formulated using the cited paper's content, providing the evidentiary basis for the claim. 
The hypothesis is tailored to assert relevance at each of the three levels (document, section, and sentence). 
This formulation is structured as follows:

\begin{tcolorbox}[
    colback=black!5!white,  
    colframe=darkgray!75!darkgray, 
    fonttitle=\bfseries,
    fontupper=\small,
    title=NLI Task Formulation
]
\noindent\textbf{Premise} (Consistent for all levels):
{There is a paper. Title: ``$T_r$''. Abstract: $A_r$.}

\noindent\textbf{Hypotheses} (Tailored for each level of granularity):
\begin{itemize}[leftmargin=*, topsep=3pt, itemsep=2pt]
    \item \textbf{Document-level:} {The paper titled ``$T_r$'' with the given abstract is thematically relevant to the survey titled: ``$T_S$''.}
    \item \textbf{Section-level:} {The paper titled ``$T_r$'' with the given abstract is relevant to the section: ``\textit{Section Title}''.}
    \item \textbf{Sentence-level:} {The paper titled ``$T_r$'' with the given abstract supports the claim made in the sentence: ``\textit{Sentence Text}''.}
\end{itemize}
\end{tcolorbox}

The score for each citation unit is calculated as follows. 
Let $R$ denote the set of all citation instances in the generated survey. For each citation $r \in R$, we compute a score at each of the three levels: document ($y_d(r)$), section ($y_s(r)$), and sentence ($y_t(r)$). 
First, we resolve two special cases without querying the NLI model. 
Any citation $r$ not found in our academic corpus is classified as a hallucination and assigned a score of $y_x(r) = 0$ at all levels $x \in \{d, s, t\}$. 
Conversely, any citation $r$ that is present in the ground-truth survey's bibliography is assigned a document-level score of $y_d(r)=1$.
For the remaining cases, the relevance score $y_x(r)$ is derived from the NLI model's output distribution. Let $p_{\text{ent}}$, $p_{\text{neu}}$, and $p_{\text{con}}$ denote the predicted probabilities for the \textsc{Entailment}, \textsc{Neutral}, and \textsc{Contradiction} labels, respectively. We map these probabilities to a discrete score as follows:

\begin{equation}
{
\begin{aligned}
y_x(r) = &\begin{cases}
1, & \text{if } p_{\text{ent}} > \max(p_{\text{neu}}, p_{\text{con}}) \\
0.5, & \text{if } p_{\text{neu}} > \max(p_{\text{ent}}, p_{\text{con}}) \\
0, & \text{otherwise.}
\end{cases} \\
&(\text{where } x \in \{d, s, t\})
\end{aligned}}
\end{equation}

Finally, we aggregate these citation scores ($y_x(r)$) for the final metrics: {Document-level Accuracy} ($R_d$), {Section-level Accuracy} ($R_s$), and {Sentence-level Accuracy} ($R_t$). 
For each level $x \in \{d, s, t\}$, the score $R_x$ is calculated as the mean of the individual citation scores $y_x(r)$ over all citation instances $\mathcal{R}$ in the survey:

{
\begin{equation}
\label{eq:ref_accuracy} 
R_x = \frac{1}{|{R}|} \sum_{r \in {R}} y_x(r), \quad x \in \{d, s, t\},
\end{equation}
}

\subsection{Structural Quality}
\label{sec:structure}

The structure of a scientific survey fundamentally determines its logical flow and coherence, making structural quality critical for readability. To evaluate this, we introduce two complementary metrics assessing the outline at both macroscopic and microscopic levels.
Our first metric, the \textbf{Structure Quality Score (SQS)}, evaluates high-level organization by comparing the structure, meaning, and wording of generated versus ground-truth headings. Complementing this, our second metric, \textbf{Soft-Heading Recall (SHR)}, provides a fine-grained evaluation of heading alignment. Specifically, it quantifies the correspondence between generated and ground-truth headings via semantic-embedding similarity.

\paragraph{Structure Quality Score (SQS)} SQS evaluates the structural fidelity of a generated survey by assessing its alignment with the ground truth. To capture both the semantic content and the logical organization of the sections, we adopt the {LLM-as-a-Judge} paradigm. Specifically, we provide the LLM with the hierarchical heading structures from both the generated and ground-truth surveys. Crucially, to preserve the structural hierarchy (e.g., section vs. subsection), we formulate these inputs using explicit markup (e.g., Markdown or LaTeX) rather than flat lists. 
The model is then prompted to assign a similarity score based on the correspondence of structure, meaning, and wording, following a comprehensive scoring rubric\footnote{The specific prompt template for the scoring rubric is available in our official repository at \url{https://github.com/oneal2000/SurGE/blob/main/src/structureFuncs.py}}.

\paragraph{Soft-Heading Recall (SHR)}
To measure fine-grained alignment, SHR evaluates how well the generated outline covers the specific headings present in the ground-truth outline. Unlike metrics based on exact lexical matching, SHR leverages semantic similarity to robustly handle variations in wording and paraphrasing.
Formally, SHR is defined as the soft cardinality overlap between the predicted heading set ($H_P$) and the ground-truth heading set ($H_{GT}$):
{
\begin{align}
\text{SHR} &= \frac{\mathcal{S}(H_P \cap H_{GT})}{\mathcal{S}(H_{GT})},
\end{align}}

\noindent where $\mathcal{S}(A)$ denotes the soft cardinality of a heading set $A$. Intuitively, this metric counts the number of semantically unique headings in a set by down-weighting redundant headings. Specifically, the contribution of each heading is inversely proportional to its aggregated similarity with all other headings in the set:
\begin{align}
\mathcal{S}(A) &= \sum_{i=1}^{K} \frac{1}{\sum_{j=1}^{K} \text{sim}(A_i, A_j)}.
\end{align}
Here, $\text{sim}(A_i, A_j)$ is the cosine similarity between the embeddings of headings $A_i$ and $A_j$.
A standard set intersection would be too strict for comparing paraphrased headings. Therefore, we define the soft intersection cardinality using the inclusion-exclusion principle:

\begin{equation}
{\small
\begin{split}
\mathcal{S}(H_P \cap H_{GT}) &= \mathcal{S}(H_P) + \mathcal{S}(H_{GT})  - \mathcal{S}(H_P \cup H_{GT}).
\end{split}
}
\end{equation}

\noindent The core idea lies in the union term, $\mathcal{S}(H_P \cup H_{GT}$.
When computed on the combined heading set, a predicted heading and a similar ground-truth heading mutually reduce the union's soft cardinality. This reduction directly quantifies their semantic overlap, allowing the metric to reward paraphrased matches. A higher SHR score thus indicates better granular alignment.

\subsection{Content Quality}
\label{sec:content}

To assess the content quality of generated scientific surveys, we propose the Content Quality Score (CQS) metric based on the {LLM-as-a-Judge} paradigm, leveraging GPT-4o to evaluate each section of the survey. 
The evaluation is based on five criteria: fluency and coherence, logical clarity, avoidance of redundancy, clarity of description, and absence of errors. 
To guide the LLM's evaluation, we designed a detailed instruction prompt for the LLM~\footnote{The detailed instruction prompt is available in our official repository at \url{https://github.com/oneal2000/SurGE/blob/main/src/informationFuncs.py}}. 
Each section is scored on a scale of 0 to 5, where a higher score reflects superior fluency, logical progression, and clarity.
Given the LLM's context-length limitations, we score each survey section independently, and the final score is the average of the section scores.

Additionally, we also report ROUGE and BLEU scores in our main experiments for completeness. 
\textit{However, both ROUGE and BLEU are provided strictly for reference and do not constitute part of the metric suite in our SurGE benchmark.} Traditional lexical matching metrics fall short in accurately measuring the performance of open-ended scientific generation, as they fail to capture the semantic correctness and logical flow essential for high-quality surveys.

\begin{table}[t]
    \centering
\caption{Human-LLM alignment meta-evaluation. We measure the correlation between human rankings and our metrics: Structure (SQS) and Content (CQS). \textbf{w/ GT} includes human-written ground truth in the ranking list, while \textbf{w/o GT} considers generated baselines only.}    \label{tab:human_alignment}
    {
    \begin{tabular}{lcccc}
        \toprule
        \multirow{2}{*}{\textbf{Metric}} & \multicolumn{2}{c}{\textbf{SQS}} & \multicolumn{2}{c}{\textbf{CQS}} \\
        \cmidrule(lr){2-3} \cmidrule(lr){4-5}
         & \textbf{w/ GT} & \textbf{w/o GT} & \textbf{w/ GT} & \textbf{w/o GT} \\
        \midrule
        Kendall's $\tau$ ($\uparrow$) & 0.805 & 0.610 & 0.797 & 0.594 \\
        Spearman's $\rho$ ($\uparrow$) & 0.883 & 0.707 & 0.878 & 0.695 \\
        \bottomrule
    \end{tabular}
    }
\end{table}

\subsection{Meta-Evaluation: Human Alignment}
\label{sec:meta_eval}

Alignment with expert consensus is a prerequisite for automated evaluation. To validate our metrics, we compared the rankings produced by SurGE against human expert judgments.
We first select a representative subset of baselines (Section~\ref{sec:baselines}): {RAG}, {StepSurvey}, and {AutoSurvey}, all utilizing the {Qwen2.5-14B-Instruct} backbone.
The human evaluation was conducted by doctoral researchers in Computer Science, all of whom possess prior experience in academic peer review.
Specifically, following a blind review protocol, for each test query, these experts ranked four anonymous surveys (the three LLM-outputs plus the human-written {Ground Truth}) and were tasked with producing two independent ranking lists based on:
(1) {Structural Quality}, adhering to the definitions in Section~\ref{sec:structure}; and 
(2) {Content Quality}, following the metrics in Section~\ref{sec:content}.
We quantified alignment using Kendall’s $\tau$~\citep{kendall1938new} and Spearman’s $\rho$~\citep{spearman1904proof}. 
As human-written surveys significantly outperform models while model-to-model differences are more subtle, we evaluate alignment in two distinct settings:
(1) \textbf{w/ GT}: calculating correlation across the full list; and
(2) \textbf{w/o GT}: calculating correlation excluding the GT to assess the {discriminative power} among competitive baselines.


Table~\ref{tab:human_alignment} presents the human-model alignment measured by Kend-all’s $\tau$ and Spearman’s $\rho$.
In the \textbf{w/ GT} setting, SurGE achieves a high $\tau$ of $ 0.805$, confirming strong alignment in distinguishing human-level scholarship from machine-generated text.
Crucially, in the more challenging \textbf{w/o GT} setting, our framework maintains a substantial correlation (Spearman's $0.707$, Kendall's $0.610$).
Prior studies in NLG evaluation suggest that these values indicate strong reliability.
For instance, meta-evaluations of summarization metrics often report Spearman correlations in the range of $0.5$ to $0.7$ as state-of-the-art performance~\citep{liu2023geval, fabbri2021summeval}.

\section{Experimental Setup}
\label{sec:baselines}
\label{sec:5}
In this section, we detail the implementation of our experiment. 
Each baseline follows a two-stage pipeline: (1)~{retrieving} a set of potentially relevant papers for a given topic, and (2)~{organizing and summarizing} the retrieved papers to produce a structured survey.

\subsection{Paper Retriever}
\label{sec:retriever_main}

For fair comparison, all baselines share the same dense retriever for the first stage. 
We employ a RoBERTa-based dual encoder~\cite{zhan2021optimizing} initialized from \texttt{roberta-base}~\cite{liu2019roberta}. 
Given a topic description $q$ and a paper abstract $d$, we encode them independently and use the final-layer \texttt{[CLS]} vectors as embeddings $h_q$ and $h_d$ (with shared encoder parameters). 
Specifically, we construct the query and document inputs as \texttt{[CLS]} $q$ \texttt{[SEP]} and \texttt{[CLS]} $d$ \texttt{[SEP]}, respectively.
We score a pair by dot product $s(q,d)=h_q^\top h_d$ and optimize the following softmax cross-entropy loss over one positive $d^+$ and negatives $N$ sampled from the corpus:
\begin{equation}
\mathcal{L}(q,d^+,N)=-\log\frac{\exp(s(q,d^+))}{\exp(s(q,d^+))+\sum_{d^- \in N}\exp(s(q,d^-))}.
\end{equation}

For each topic, the positive papers are derived from the ground-truth relevant papers in the benchmark, while negative papers are randomly sampled from the remaining corpus.
This objective encourages the retriever to assign higher scores to relevant papers than to irrelevant ones.
After training, given a topic description, the retriever scores candidate papers in the corpus and returns the top-ranked papers as the input evidence for the downstream generation stage.

\subsection{Baselines}\label{sec:baselines}

We evaluate the following four LLM-based survey generation methods. For fair comparison, all baselines use the same Paper Retriever to collect the top-100 candidate papers for topic $t$. More implementation details are provided in our official GitHub repository.

\begin{itemize}[leftmargin=*,itemsep=0pt]
\item \textbf{Retrieval-Augmented Generation}~\cite{lewis2020retrieval}. 
We retrieve the top-100 papers for topic $t$ and split them into batches to fit the context window.
An LLM first summarizes each batch with paper-ID citations preserved, and a second LLM call fuses these partial summaries into a coherent survey.

\item \textbf{AutoSurvey}~\citep{wang2024autosurvey} employs an outline-expand-refine pipeline.
We follow its original stages but replace its retriever with ours and keep the same top-100 candidate pool.
The outline is expanded section-by-section using the papers most relevant to each section, followed by a final consistency and citation-format refinement before merging sections.

\item \textbf{StepSurvey}~\citep{lai2024instruct} adopts a step-by-step plan-and-write process.
Given the top-100 retrieved papers, it first proposes a title and primary headings, then derives finer-grained subtopics and associates each with a relevant paper subset.
The LLM drafts subsections sequentially under this plan.

\item \textbf{SurveyForge}~\citep{yan2025surveyforge} introduces a two-stage pipeline with heuristic outlining and memory-driven writing.
We follow the official implementation\footnote{\url{https://github.com/InternScience/SurveyForge}} but replace its retrieval component with our retriever and keep the candidate pool size the same.
It builds a hierarchical outline and uses its Scholar Navigation Agent to draft each subsection grounded on the retrieved papers.
\end{itemize}





\subsection{Implementation Details}\label{sec: Implementation Details}
\textbf{Retriever Training.} For the training of the Paper Retriever, we randomly split the dataset into training and test sets at a ratio of 4:1. The retriever is initialized using the pre-trained RoBERTa model~\citep{liu2019roberta}. We adopt the AdamW optimizer with a learning rate of $5\times10^{-6}$, and the model is trained for $10$ epochs using mixed-precision (fp16) training. At inference time, each query retrieves the top-100 relevant papers based on similarity scores.

\noindent \textbf{Model Configuration.} To ensure a fair comparison among the pipeline-based baseline methods ({AutoSurvey}, {StepSurvey}, and {SurveyForge}), which originally utilize different underlying models, we unify their base LLM to {Qwen2.5-14B-Instruct}~\citep{yang2024qwen2}. 
For the {Retrieval-Augmented Generation (RAG)} baseline, we extend our evaluation to include three distinct backbone models to analyze performance across different model capabilities: {Qwen2.5-14B-Instruct}, {GPT-4o}~\cite{hurst2024gpt}, and {Gemini-2.5-Pro-Thinking}~\cite{comanici2025gemini}.
For the generation configuration of open-source models, we utilize the implementation provided by Hugging Face with default hyperparameters and the official chat template\footnote{\url{https://huggingface.co/Qwen/Qwen2.5-14B-Instruct}}. 
For the embedding model of the SHR metric, we use the \texttt{bge-large-en-v1.5} model~\cite{chen2024bge}.
All experiments are conducted on a GPU server equipped with 1TB RAM and eight NVIDIA A100 GPUs (40GB memory each).

\begin{table}[t]
\centering

\caption{Comparison of retrieval models on recalling ground-truth cited papers. R@k denotes recall within the top $k$ retrieved documents. The best results are in bold.}
\label{tab:retrieval_recall}

\resizebox{1\columnwidth}{!}{
\begin{tabular}{lcccccc}
\toprule
\textbf{Model} & \textbf{R@20} & \textbf{R@30} & \textbf{R@100} & \textbf{R@200} & \textbf{R@500} & \textbf{R@1000} \\
\midrule
\textbf{BM25} & 0.0548 & 0.0652 & 0.1193 & 0.1596 & 0.2213 & 0.2715 \\
\textbf{PR} & \textbf{0.1706} & \textbf{0.2145} & \textbf{0.3665} & \textbf{0.4681} & \textbf{0.6011} & \textbf{0.6805} \\
\bottomrule
\end{tabular}

}
\end{table}

\begin{table*}[t]
\centering
\caption{Experimental results comparing survey generation baselines across four dimensions: Comprehensiveness (Comp.), Citation Accuracy, Structural Quality, and Content Quality. Metrics include Recall, Document/Section/Sentence-level Citation Accuracy (Doc-Acc, Sec-Acc, Sent-Acc), Structure Quality Score (SQS), Soft-Heading Recall (SHR), ROUGE-L, BLEU, and Content Quality Score (CQS). The best results within each group are in \textbf{bold}. For fair comparison, all specialized survey generation agents are implemented using the same base LLM, Qwen-2.5-14B.}

\label{table:main}
{
\begin{tabular}{lccccccccc}
\toprule
                     & \multicolumn{1}{c}{\textbf{Comp.}} & \multicolumn{3}{c}{\textbf{Citation Accuracy}} & \multicolumn{2}{c}{\textbf{Structural Quality}} &\multicolumn{3}{c}{\textbf{Content Quality}}\\
\cmidrule(lr){2-2}\cmidrule(lr){3-5}\cmidrule(lr){6-7}\cmidrule(lr){8-10}
\textbf{Baseline}    & \textbf{Recall} & \textbf{Doc-Acc} & \textbf{Sec-Acc} & \textbf{Sent-Acc} & \textbf{SQS} & \textbf{SHR} & \textbf{R-L} & \textbf{BLEU} & \textbf{CQS} \\
\toprule
\multicolumn{10}{c}{\textit{General-Purpose RAG Baselines}} \\
\midrule

\textbf{RAG-Qwen}         & 0.0214 & 0.2857 & 0.2502 & 0.2500 & 0.6829 & \textbf{0.7900}   & \textbf{0.1519} & \textbf{10.38} & 4.6723 \\
\textbf{RAG-GPT}     & 0.0419 & 0.4525 & 0.3166 & 0.3226 & 0.7561 & 0.6659   & 0.1409 & 8.36  & \textbf{4.9250} \\
\textbf{RAG-Gemini}  & \textbf{0.0784} & \textbf{0.4680} & \textbf{0.4056} &\textbf{ 0.4124} & \textbf{1.2927} & {0.7591}   & {0.1442} & {8.37}  & 4.5619 \\
\midrule
\multicolumn{10}{c}{\textit{Specialized Survey Generation Agents}} \\
\midrule
\textbf{AutoSurvey}  & 0.0351 & 0.3617 & \textbf{0.4935} & \textbf{0.4870} & \textbf{1.3902} & {0.9697} & {0.1578} & {10.44} & 4.7390 \\
\textbf{StepSurvey}  & 0.0630 & 0.4576 & 0.4571 & 0.4636 & 1.1951 & \textbf{0.9763} & \textbf{0.1590} & \textbf{12.02} & \textbf{4.8451} \\
\textbf{SurveyForge} & \textbf{0.0868} & \textbf{0.4719} & {0.4651} & {0.4772} & {1.0537} & 0.9493   & 0.1443 & 8.99  & 4.8070 \\
\bottomrule
\end{tabular}
}

\label{tab:combined_results}
\end{table*}

\section{Experimental Results}\label{sec:exp_results}

This section presents the experimental results. We begin by analyzing the performance of the Paper Retriever (\S\ref{sec:retrieval_analysis}), then analysis the end-to-end survey generation performance of the baselines (\S\ref{sec:overall_performance}).

\subsection{Evaluation of the Retrieval Stage}
\label{sec:retrieval_analysis}

A crucial question in our two-stage pipeline is whether the low reference recall of baseline systems stems from the retriever's inability to find relevant papers (a data coverage issue) or the generator's inability to identify them (a capability issue). To disentangle these factors and validate the quality of our constructed corpus, we evaluate the performance of our fine-tuned dense retriever in isolation.

We compare our dense retriever (PR) against the lexical baseline BM25~\citep{robertson2009probabilistic}, using Recall@k. As shown in Table~\ref{tab:retrieval_recall}, our fine-tuned Paper Retriever substantially outperforms the BM25 baseline. More importantly, the results provide strong evidence for the validity of our benchmark data. At $k=1000$, the retriever successfully recalls 68.05\% of the ground-truth references. This high upper-bound coverage confirms that the vast majority of expert-selected references are indeed present and retrievable within our corpus, addressing potential concerns regarding the completeness of the data collection process.
Furthermore, the substantial disparity between the retrieval upper bound (68.05\%) and end-to-end baseline performance ($<10\%$) indicates that the bottleneck of survey generation lies in the agents' capacity to select and synthesize information. While this confirms the value of SurGE as a challenging testbed, the retrieval stage itself presents further opportunities for optimization. Future research could investigate sophisticated retrieval paradigms such as task-oriented search agents.

\subsection{Evaluation of End-to-End Generation}\label{sec:overall_performance}
To empirically investigate the capabilities of current survey generation systems, we benchmark a diverse set of approaches on the SurGE dataset.
Table~\ref{table:main} presents the comparative results across four critical dimensions. Our analysis reveals four key observations regarding the trade-offs between retrieval capabilities, structural coherence, and content fidelity:

\begin{itemize}[leftmargin=*]

\item  \textbf{Surface-level fluency does not imply factual accuracy.}
We observe that RAG-Qwen achieves competitive scores on n-gram metrics, including ROUGE-L and BLEU. However, it suffers from the lowest Citation Accuracy and Recall. This disparity highlights a "fluency hallucination" trap, in which naive RAG models can generate coherent yet factually ungrounded text. This underscores the need to rely on fact-centric metrics (e.g., Doc-Acc, CQS) rather than traditional n-gram metrics for evaluating the quality of scientific survey generation.

\item  \textbf{Agentic workflows are essential for structural quality.}
All specialized agents achieve higher Soft-Heading Recall (SHR) than standard RAG baselines, showing that explicit planning and structured drafting are important for survey generation. 
For the Structure Quality Score (SQS) metric, AutoSurvey achieves the best overall performance, while other agentic designs remain competitive. Overall, these results support the need for explicit planning stages to reliably capture hierarchical discourse structures in academic surveys.

\item  \textbf{Specialized mechanisms compensate for limited model capacity.}
Comparing RAG variants demonstrates that the base LLM's general capabilities are important, as RAG-Gemini significantly outperforms RAG-Qwen. However, SurveyForge reverses this reliance on scale: by using Qwen-14B, it not only outperforms its direct counterpart, RAG-Qwen, but also surpasses RAG-Gemini in Citation Accuracy. 
This suggests that a well-structured agentic framework can compensate for limited model capacity by decomposing survey generation into more controllable substeps, such as literature retrieval, evidence organization, citation selection, and section-level planning, thereby enabling smaller models to produce more faithful and citation-grounded outputs.

\item  \textbf{Different agentic frameworks yield divergent strengths.}
While SurveyForge excels at retrieving accurate literature (Recall and Doc-Acc), StepSurvey dominates in content-quality metrics, achieving the highest CQS. This indicates that SurveyForge's memory-driven mechanism ensures better coverage of relevant papers, whereas StepSurvey's granular, step-by-step drafting process yields superior linguistic quality.

\end{itemize}

\section{Related Work}

\subsection{Survey Generation}

Before the recent advances in LLMs, scientific survey generation was commonly studied as an advanced form of Multi-Document Summarization (MDS), in which a system is expected to synthesize information from multiple papers into an overview~\citep{lu2020multi,ma2022multi}. 
However, survey generation goes beyond conventional MDS, as it requires not only concise information aggregation but also large-scale literature organization, cross-paper synthesis, and hierarchical discourse construction~\citep{nayeem2024,bleiweiss2023,wang2024autosurvey}. 
These requirements made the task particularly challenging in the pre-LLM era, as earlier models were limited in long-context modeling and global planning, making it difficult to produce lengthy, well-structured surveys with consistent factual grounding.

The emergence of LLMs has substantially expanded the feasibility of automated survey generation, shifting the task from short-form summarization toward the construction of full-length academic manuscripts. 
Recent studies have begun to explore this direction through agentic pipelines that decompose survey writing into multiple stages, such as literature collection, outline construction, section-level drafting, and iterative refinement~\citep{liang2025surveyx}. 
For instance, AutoSurvey~\citep{wang2024autosurvey} and StepSurvey~\citep{lai2024instruct} address the long-context challenge via hierarchical decomposition, employing outline-driven or step-by-step strategies to iteratively refine content from headings to subsections. 
To enhance information utilization, SurveyForge~\citep{yan2025surveyforge} integrates a Scholar Navigation Agent (SANA) with memory-driven writing mechanisms.
While these works demonstrate the feasibility of automated survey generation, they primarily focus on pipeline construction rather than standardized evaluation.

\subsection{Retrieval-Augmented Generation}

Retrieval-Augmented Generation (RAG) has become a widely adopted paradigm for enhancing large language models (LLMs) with external knowledge, where task-relevant information is retrieved from external sources and used to ground model generation~\cite{lewis2020retrieval,dong2025decoupling,tu2025robust,su2025dynamic,tu2026analytical,su2025sigirap}. 
Prior studies have demonstrated the benefits of RAG in several important scenarios, including mitigating hallucinations~\cite{wang2026joint,su2025towards,su2024unsupervised}, enabling timely knowledge updates~\cite{wang2025decoupling,wang2025knowledge,wang2025decoupling}, and facilitating domain adaptation without requiring expensive full-parameter retraining~\cite{su2025judge,su2024stard,wang2024lekube}. 
A typical RAG framework adopts a retrieve-and-generate workflow: given an input query, the system first identifies relevant documents from a large-scale corpus~\cite{robertson2009probabilistic,su2024wikiformer,fang2024scaling,su2025pre,tu2026generalized}, and then conditions the LLM on the retrieved evidence to produce a grounded response. 
Building on this basic pipeline, recent research has further extended RAG along multiple directions, such as dynamic RAG~\cite{jiang2023active,su2024dragin,su2024mitigating}, graph RAG~\cite{edge2024local}, parametric RAG~\cite{su2025parametric,tan2025dynamic}, and agentic RAG~\cite{jin2025search,su2026skill}.

From a broader perspective, survey generation intersects with Retrieval-Augmented Generation (RAG) and long-context processing, yet remains distinct. Unlike the standard retrieve-then-read paradigm of RAG, surveys require complex hierarchical planning and multi-paper synthesis. Similarly, while benchmarks like LongBench~\citep{bai2024longbench} and L-Eval~\citep{an2024eval} assess long-input processing, they fail to capture survey-specific requirements, such as citation coverage and claim-to-citation faithfulness.
Consequently, a standardized evaluation protocol remains absent, as existing assessments rely heavily on unscalable human judgments or biased self-validation protocols. SurGE addresses this critical gap by establishing the first method-agnostic benchmark for the survey generation task.

\section{Conclusion}
\label{sec:6}
In this paper, we introduce \textbf{SurGE}, a benchmark and evaluation framework designed to address the critical need for standardized and reproducible evaluation in automated scientific survey generation. 
SurGE provides a large-scale academic corpus, a set of expert-written ground-truth surveys, and a fully automated framework to evaluate surveys on comprehensiveness, citation accuracy, structure, and content.
Our experiments reveal the limitations in SOTA LLM-based survey generation systems, highlighting challenges such as incomplete topic coverage and reference hallucination. 
We believe SurGE paves the way for developing more trustworthy and effective systems for survey generation.

\bibliographystyle{ACM-Reference-Format}
\balance
\bibliography{sample-base}

\end{document}